\documentclass[letterpaper, 10 pt, conference]{ieeeconf}

\IEEEoverridecommandlockouts                              

\usepackage[utf8]{inputenc} 
\usepackage[T1]{fontenc}  
\usepackage{hyperref}     
\usepackage{url}      
\usepackage{booktabs}     
\usepackage{amsfonts}     
\usepackage{nicefrac}     
\usepackage{microtype}    

\usepackage{amsmath,amssymb}
\usepackage[usenames,dvipsnames,svgnames,table]{xcolor}
\usepackage{graphicx}
\usepackage{wrapfig}
\usepackage{stmaryrd}
\usepackage{xcolor}

\usepackage{subcaption}
\usepackage{listings} 

\definecolor{darkred}{rgb}{.7,0,0}
\definecolor{darkgreen}{rgb}{0,.5,0}
\definecolor{darkblue}{rgb}{0,0,.8}
\definecolor{darkcyan}{rgb}{0,0.6,.6}
\definecolor{darkorange}{rgb}{.8,.4,0}
\definecolor{gray}{rgb}{.4,.4,.4}

\DeclareMathOperator*{\argmin}{\text{argmin}}

\newcommand{\node}{n}
\newcommand{\children}{\mathcal{C}}
\newcommand{\rootnode}{\node_0}

\newcommand{\state}{s}
\newcommand{\stateset}{\mathcal{S}}
\newcommand{\goalset}{\mathcal G} 

\newcommand{\initialstate}{\mathcal I}
\newcommand{\streams}{ \Psi}
\newcommand{\predicates}{\mathcal P}
\newcommand{\actions}{\mathcal{A}}
\newcommand{\lifted}{\texttt{LAZY}}

\newcommand{\pol}{\pi}
\newcommand{\depthz}{d_0}  
\newcommand{\search}{search}

\newcommand{\trans}{T}   

\newcommand{\fringe}{\mathcal{Q}}

\newcommand{\closed}{\text{closed}}

\newcommand{\sample}{\text{refine}}
\newcommand{\groundings}{\vec{\theta}}
\newcommand{\hlplan}{\vec{a}}
\newcommand{\stream}{\text{stream}}
\newcommand{\stats}{\Phi}
\newcommand{\priority}{f}

\newcommand{\feasibility}{\phi}

\makeatletter
\let\NAT@parse\undefined
\makeatother

\newcommand\mydots{\hbox to 1em{.\hss.\hss.}}

\begin{document}

\title{\LARGE \bf
Policy-Guided Lazy Search with Feedback for Task and Motion Planning
}

\author{
Mohamed Khodeir$^{1}$
\quad
Atharv Sonwane$^{2}$
\quad 
Ruthrash Hari$^{3}$
\quad 
Florian Shkurti$^{1}$
\thanks{$^{1}$Robot Vision and Learning Lab, University of Toronto Robotics Institute.}
\thanks{$^{2}$Dept. of Computer Science, University of BITS Pilani}
\thanks{$^{3}$Dept. of Mathematical and Computational Sciences, University of Toronto}
\thanks{
    {\tt\small \{m.khodeir, ruthrath.hari\}@mail.utoronto.ca, atharvs.twm@gmail.com, florian@cs.toronto.edu}
}
}

\maketitle


\IEEEpeerreviewmaketitle

\begin{abstract}
PDDLStream solvers have recently emerged as viable solutions for Task and Motion Planning (TAMP) problems, extending PDDL to problems with continuous action spaces. 
Prior work has shown how PDDLStream problems can be reduced to a sequence of PDDL planning problems, which can then be solved using off-the-shelf planners. However, this approach can suffer from long runtimes. In this paper we propose $\lifted$, a solver for PDDLStream problems that maintains a single integrated search over action skeletons, which gets progressively more geometrically informed, as samples of possible motions are lazily drawn during motion planning. We explore how learned models of goal-directed policies and current motion sampling data can be incorporated in $\lifted$ to adaptively guide the task planner. We show that this leads to significant speed-ups in the search for a feasible solution evaluated over unseen test environments of varying numbers of objects, goals, and initial conditions. We evaluate our TAMP approach by comparing to existing solvers for PDDLStream problems on a range of simulated 7DoF rearrangement/manipulation problems. Code can be found at \url{https://rvl.cs.toronto.edu/learning-based-tamp}.
\end{abstract}
\vspace{-0.2cm}


\section{Introduction}
\label{sec:intro}
Task and motion planning (TAMP) problems are challenging because they require reasoning about both discrete and continuous decisions that are interdependent. TAMP solvers typically decompose the problem by using a symbolic task planner that searches over discrete abstract actions, such as which object to interact with or what operations are applicable, and a motion planner which attempts to find the continuous parameters that ground those abstract actions, for instance grasp poses and robot configurations. The motion planner informs the task planner when backtracking is necessary. Thus, the interplay between abstract task planning and low-level motion planning has a significant effect on both runtime and percentage of problems solved.


In this work, we provide a significantly improved PDDLStream~\cite{garrett2020pddlstream} solver ($\lifted$) for task and motion planning problems, which learns to plan from experience and adapts based on current execution data. The motion planner of our solver provides feasibility updates to a priority/guidance function that is used to inform action selection by the symbolic task planner. $\lifted$ plans optimistically and lazily (deferring motion sampling until an action skeleton is found), and maintains a single unified search tree, as opposed to solving a sequence of PDDL problems over a growing set of facts, as was done in \cite{garrett2020pddlstream} and its current variants. 
\newpage

\begin{figure}[t]
     \begin{subfigure}[b]{0.45\linewidth}
         \centering
         \includegraphics[width=\textwidth]{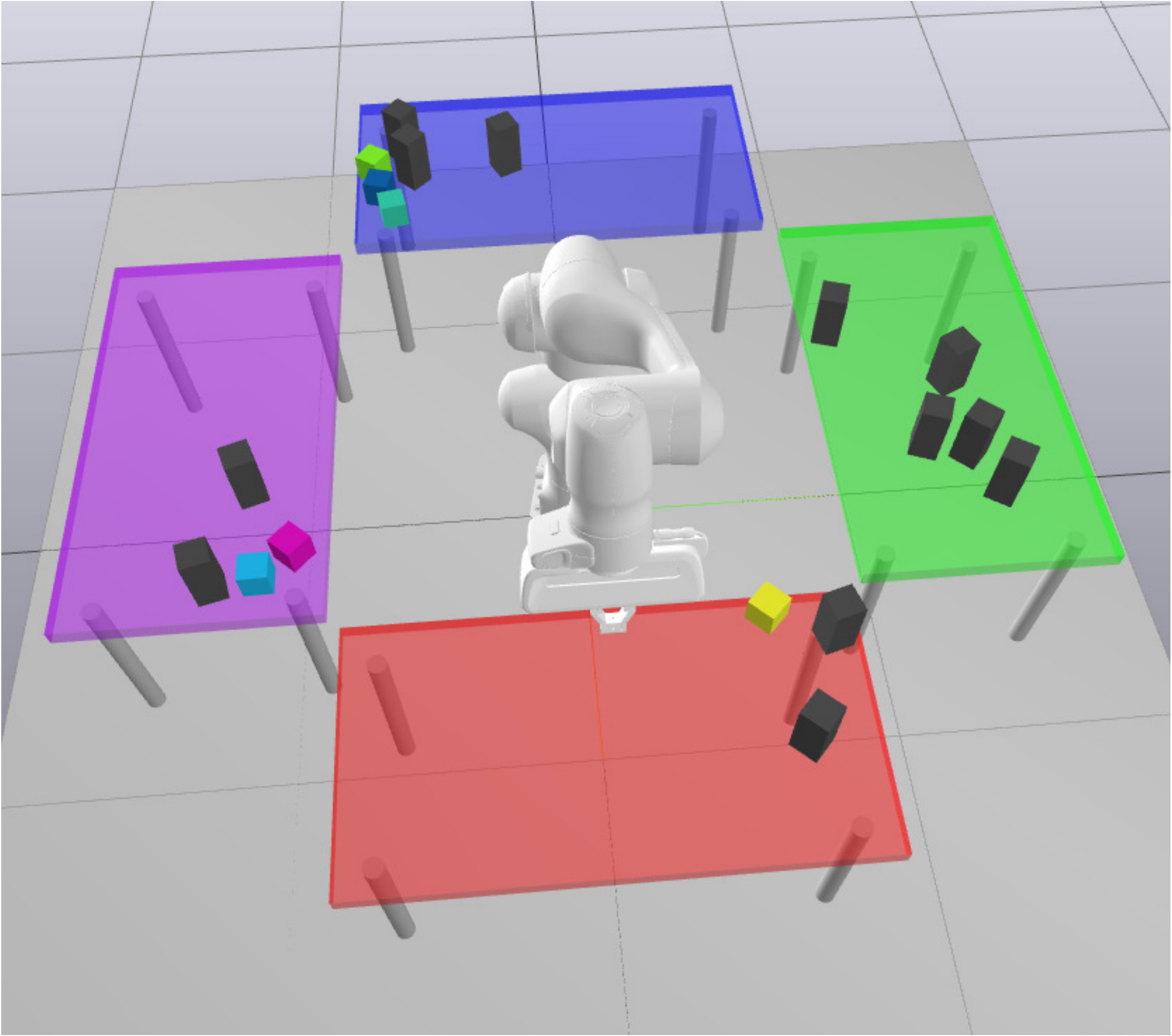}
         \label{fig:clutter}
     \end{subfigure}
     \begin{subfigure}[b]{0.45\linewidth}
         \centering
         \includegraphics[width=\textwidth]{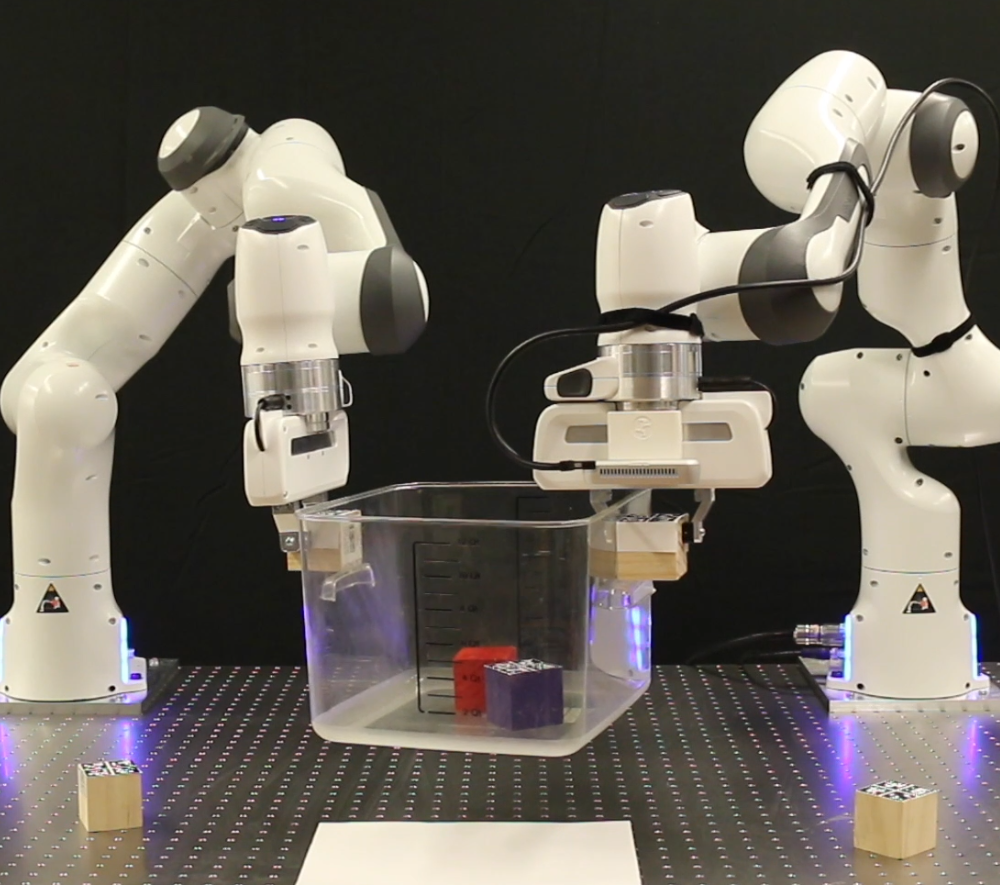}
         \label{fig:distractors}

     \end{subfigure}
     
     \begin{subfigure}[b]{\linewidth}
     \vspace{-0.8cm}
         \centering
         \includegraphics[width=\textwidth]{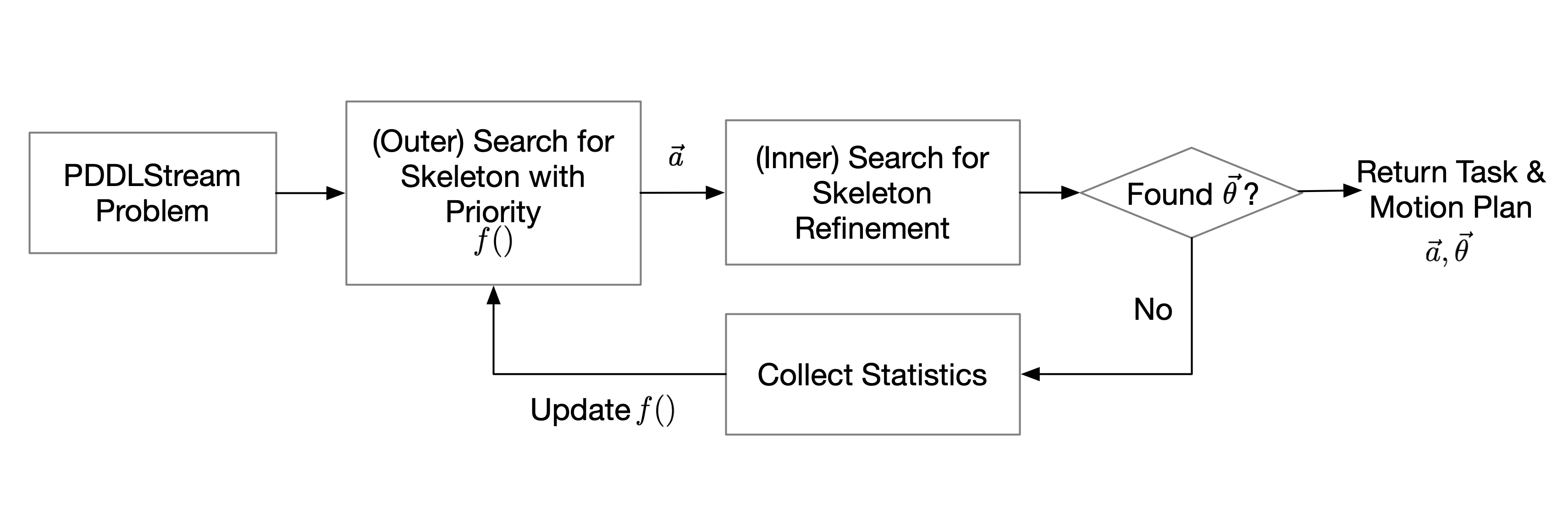}
     \end{subfigure}
     \vspace{-0.8cm}
    \caption{Top Left: Simulated evaluation tasks in Clutter environment. Top Right: Real-world manipulation problems using two 7DoF robot arms. Bottom: A flowchart illustrating the high level components of our approach. }
    \label{fig:experiments}
\end{figure}
\unskip
\vspace*{-1.225cm}
A core component of our method is a goal-conditioned policy over high-level actions, which we learn using behaviour cloning on past planning demonstrations. This policy is treated as a priority function, which guides the action skeleton search performed by the task planner towards promising abstract action sequences. While this can often eliminate the need for backtracking altogether, the policy may still predict geometrically infeasible actions in more challenging TAMP problems. We therefore show how the predictions of this priority function can be updated online in response to failed samples in motion planning, allowing successive iterations of the task planner to focus the search on more feasible action sequences. 
The result is a policy-guided bi-level search for TAMP problems, which improves online from experience and past data, and demonstrates impressive planning performance on unseen environments from a test distribution, while being trained with only a few hundred demonstrations. 

Our main contributions are: (1) A lazy search framework for PDDLStream problems, which maintains a single search tree over symbolic plan skeletons. (2) A method for incorporating a learned policy over symbolic actions into sampling-based bi-level search, and efficiently updating it online using feedback from motion planning. (3) A concrete parametrization of this goal conditioned policy as a Graph Attention Network (GAT) which incorporates both high-level and low-level state. We empirically evaluate our proposed method compared to existing approaches for sampling-based TAMP, such as \cite{garrett2020pddlstream} and show significant (37\%) improvement in the number of unseen problems solved within the allotted planning time.





\section{Background: PDDL and PDDLStream}
\label{sec:Setup}
We adopt PDDLStream~\cite{garrett2020pddlstream} as the formalism for expressing TAMP problems. A PDDLStream domain $(\predicates, \actions, \streams)$ is defined by predicates $\predicates$, actions $\actions$, and streams $\streams$.

 At a high level, predicates are boolean valued n-ary functions which indicate the presence of particular relations among their variables. For instance, the predicate \texttt{isOn} may indicate that the object in its first argument is on top of the object in its second. When a predicate is applied to specific objects (e.g. \texttt{isOn(A, B)}), we refer to it as a ``fact''. 

Actions define the legal state transitions in the planning problem. They are defined by a set of parameters, a set of preconditions which define facts on those parameters which must hold in order for the action to be applicable, and effects that determine which facts about the parameters are added or removed following the application of the action.

The set of streams, $\streams$, distinguishes a PDDLStream domain from traditional PDDL. Streams are conditional generators which yield objects that satisfy specific constraints conditioned on their inputs. Formally, a stream, $s$, is defined by input and output parameters $\bar x$, $\bar o$, a set of facts $domain(s)$, and a set of facts $certified(s)$. $domain(s)$ is the set of facts that must evaluate to true for an input tuple $\bar{x}$ to be valid. This ensures the correct types of objects (e.g., configurations, poses etc.) are provided to the generators. $certified(s)$ are facts about $\bar x$ and $\bar o$ that will be true of any outputs $\bar o$ that the stream generators produce. Streams can be applied recursively to generate a potentially infinite set of objects and their associated facts, starting from those in $\initialstate$. They can also be thought of as declaratively specifying constraints between their inputs and outputs. Finally, each stream comes with a black-box procedure which, given input values $\bar x$, produces samples $\bar o$ which satisfy those constraints. We use the term stream \textit{evaluation} to refer to the act of querying this sampler.

The PDDLStream domain $(\predicates, \actions, \streams)$ defines a language in which to pose specific problems. An instance of a planning problem in this domain is defined by specifying the initial state $\initialstate$ which is simply a set of facts using predicates $\predicates$ that describe the initial scene, and the goal $\goalset$. $\initialstate$ and $\goalset$ implicitly define a set of initial objects over which facts in those sets are stated. A solution to a problem instance consists of a sequence of action instances which result in a state in which $\goalset$ is satisfied. Note that many of the parameters in a solution may need to be produced using the streams and initial objects.

Predicates in classical PDDL problems can be classified as either ``static'' or ``fluent'' depending on whether they appear in the effects of any action. Static predicates are used to define types (e.g. \texttt{isTable(x)}) or immutable relations between objects (e.g. \texttt{isSmaller(x, y)}). Fluent predicates, on the other hand, are those which can be changed by actions (e.g. \texttt{isOn(x, y)}). By definition, streams are only allowed to certify ``static'' predicates (e.g. \texttt{isGraspPose(x)}). Therefore, in PDDLStream problems, we can further categorize static predicates based on whether they are produced by streams or are simply given in the initial conditions $\initialstate$. We call the former stream-certified preconditions. 

We use the notation $\hlplan$ to refer to an ``action skeleton'', which is a sequence of discrete, high-level action instances with continuous parameters left as variables (for instance, grasp poses and placement poses).  See Fig. \ref{fig:computationgraph} for an example of a two-step action skeleton. We denote a specific assignment/grounding of continuous parameters as $\groundings$, and refer to the grounded plan as $\hlplan(\groundings)$. Similarly, we use $a$, $\theta$ and $a(\theta)$ to refer to individual actions and their grounding.

\section{Our Approach}

\begin{figure*}
    \centering
    \includegraphics[width=\linewidth]{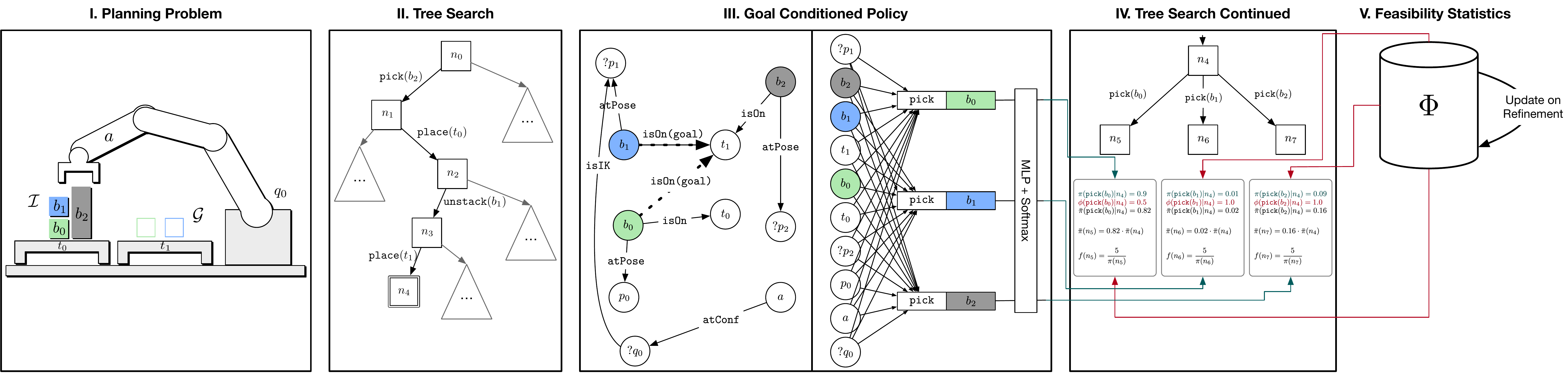}
    \caption{[I] A 2D depiction of a planning problem. Three blocks are initially placed on the table on the left side ($t_0$). The blue and green blocks ($b_0, b_1$) must be unstacked and moved to the table on the right ($t_1$), however a tall grey block ($b_2$) obstructs any grasp. [II] A snapshot of tree search where the node being expanded ($n_4$) corresponds to a partial skeleton which first moves $b_2$ and then  $b_1$ to $t_1$. The policy is now queried to determine which action to explore next. [III] The state corresponding to node $n_4$ is encoded as a graph and passed to a GAT which produces a contextual embedding of each object. A second GAT produces an embedding of the applicable actions, and the result is passed to an output layer, which computes a softmax. [IV] The tree search continues with node priorities of $n_4$'s children having been computed using the policy and empirical action feasibility estimates from the database $\Phi$. [V] When an action skeleton is found, and a refinement attempt fails, $\Phi$ is updated, leading to new priorities in subsequent tree search iterations.} 
    \label{fig:model_flow}
    \vspace{-1.5em}
\end{figure*}

\label{sec:Approach}
\subsection{Lazy Bi-Level Search}
Our overall framework is a bi-level search, similar to prior work on task and motion planning (\cite{garrett2020pddlstream}, \cite{dantam2018incremental}). In every iteration, we search for an action skeleton $\hlplan$. This outer search for an action skeleton is guided by a priority function $\priority$, which assigns a lower value to more desirable actions. We describe possible choices for how $\priority$ is defined in section \ref{sec:guidance} and elaborate on the details of skeleton search in section \ref{sec:skeletonsearch}.

Once an action skeleton is found in the outer search, we perform the inner search for grounding its continuous parameters $\groundings$. We refer to this as \textit{skeleton refinement}, and elaborate on it in section \ref{sec:refinement}.
The overall procedure terminates when refinement is successful, in which case a complete trajectory is returned. Otherwise, the result of the previous refinement is used to update the priority function $\priority$, and the next iteration begins, yielding a potentially different action skeleton. We refer to the process of incorporating the result of refinement into the priority function used by the outer search as \textit{feedback} and detail a number of possible implementations in section \ref{sec:feedback}. The search fails to solve a given problem if the allotted planning time runs out before a trajectory is found. This overall framework is summarized in Algorithm \ref{lst:lazysp} and illustrated in Figure \ref{fig:experiments}.

\begin{figure}
\vspace{-1em}
\begin{lstlisting}[label=lst:lazysp,caption={Lazy Bi-Level Search},basicstyle=\small]
def $\lifted(\rootnode$, $\goalset$, $\search$, $\priority$)
  $\stats = \emptyset$ # feasibility statistics
  while not timed out
    # $\hlplan$ is an action skeleton that achieves $\goalset$
    $\hlplan := \search(\rootnode, \goalset, \priority)$
    if $\hlplan$ = null
        break
    # maintain fail/success counts in $\stats$
    $\groundings$ := $\sample(\hlplan, N_{max}, \stats)$ 
    if $\groundings \neq null$
        # actions and their grounded parameters
        return $\hlplan(\groundings)$ 
    # some step in the plan failed
    # update priority function $\priority$
    use $\stats$ to update $\priority$
  return null # failure due to timeout
\end{lstlisting}
\vspace{-2em}
\end{figure}

\subsection{Skeleton Search Routines and their Priority Functions}
\label{sec:guidance}
There are many possible choices for the skeleton \texttt{search} routine and its associated priority function $\priority$ leading to algorithms with different characteristics. In this work, we consider two implementations of \texttt{search}: the first is a simple best-first search and the second is a beam search. Intuitively, decreasing the value of the beam width parameter in beam search allows us to create greedier search algorithms at the cost of potentially pruning out solution branches.

We also consider two implementations of $\priority$: first, the familiar A* priority function ($ \priority(\node) = g(\node) + h(\node)$), which we use to incorporate off-the-shelf domain-agnostic heuristics from prior work \cite{BONET20015}. Note that this option allows our algorithm to work well without a learned policy, using existing domain-agnostic search heuristics in place of $h$. We make use of this for data collection, and as a baseline in evaluation.

Second, we build on ideas from Levin Tree Search (LevinTS)~\cite{levints} as a way to incorporate a policy to guide the search while maintaining guarantees about completeness and search effort of the symbolic planner that relate to the quality of the policy. We assume that we are given a policy $\pol(a|\state,\goalset)$ which predicts a probability distribution over applicable discrete actions (i.e. logical state transitions) conditioned on a logical state $\state \in \stateset$ and goal $\goalset \subset \stateset$, where $\stateset$ is the set of all logical states.

We distinguish between a state in the search space and a node in the search tree by using the symbol $\state$ to denote the former and $\node$ to denote the latter. A node corresponds to a specific sequence of actions starting from the initial state $\initialstate$. We use $\rootnode$ to refer to the root node of the search tree, which is the empty path starting from the initial state $\initialstate$. Given a node $\node$ and its corresponding state and action sequence $\state_0, ..., \state_k$ $a_0, ..., a_{k-1}$, we use $\pol(a|\state,\goalset)$ to define
\begin{equation}
    \label{eq:levin}
    \pol(\node) = \prod_{i = 0}^{k-1} \pol(a_i|\state_{i},\goalset)
\end{equation}
The LevinTS priority function $f(n)$ depends on $\pol(\node)$ and the length of the sequence leading up to $\node$, which we denote $\depthz(\node)$, and is defined by:
\begin{equation}
    \label{eq:levints}
     \priority(\node) = \frac{\depthz(\node)}{\pol(\node)}
\end{equation}
LevinTS prioritizes nodes $n$ with low $f$ value, namely nodes with high probability under Eqn. \ref{eq:levin} and reachable by fewer actions than other leaf nodes in the search tree. 

\subsection{Lazy Stream Instantiation in Skeleton Search}
\label{sec:skeletonsearch}
Our outer search for an action skeleton is ``lazy'' in two respects. First, it is lazy in that it defers invoking any stream samplers until a full action skeleton has been found. In this respect, it is identical to that of the ``optimistic'' variants of PDDLStream algorithms described in prior work \cite{garrett2020pddlstream}. Second, the outer search is ``lazy'' in that streams are instantiated just-in-time, to support node expansion, as opposed to being eagerly instantiated in batch as in prior work. The main advantages of doing this are that (1) it allows a goal-seeking heuristic to guide the instantiation of streams and (2) it avoids the cost of the exhaustive search, which is incurred by prior works when eagerly instantiated streams are insufficient.

In order to implement lazy stream instantiation, we modify the logical successor function used in the tree search to determine which streams need to be instantiated in order to produce the stream-certified preconditions of a logically applicable action. 
To do this, we need to check whether all stream-certified preconditions could be produced by a combination of the set of streams $\streams$ and the objects in the current state.
We implement this check by casting it as a planning problem, where each stream defines a corresponding action with domain conditions playing the role of action preconditions, and certified conditions playing the role of action effects. We then solve for a sequence of streams that convert the initial state into a state where the desired stream-certified conditions hold. 
We use a simplified partial order planner to solve this problem for each applicable action.

\begin{figure}
    \centering
    \includegraphics[width=0.8\linewidth]{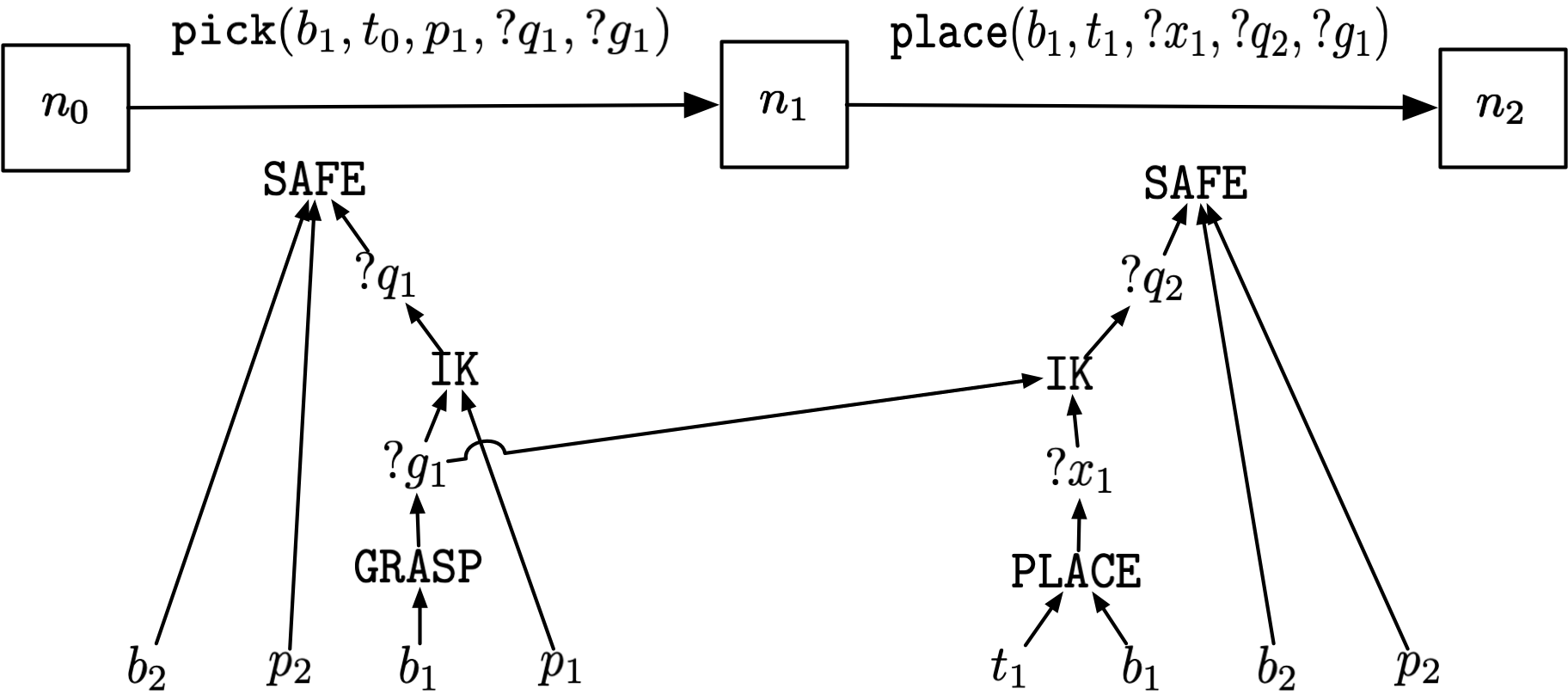}
    \caption{A plan skeleton to move an object $b_1$ from table $t_0$ to table $t_1$. There are four parameters which are unspecified: the grasp pose $?g_1$, its associated robot configuration $?q_1$, the placement pose $?x_1$ and its configuration $?q_2$. }
    \label{fig:computationgraph}
    \vspace{-1.5em}
\end{figure}
A byproduct of performing this check is that we construct a ``computation graph'' (CG) for the parameters in our plan skeleton. This takes the form of a directed acyclic hypergraph where the root nodes are objects in $\initialstate$, the hyper-edges correspond to streams, and the internal/leaf nodes are objects sampled from those streams. The CG for a given action skeleton defines a partial order over sampling operations for producing satisfying assignments of the skeleton's parameters. We maintain this structure at each node of the search tree. We refer to the stream instances that comprise the CG of a high level action $a$ as $a.streams$.  An example of a CG for a two-step action skeleton is depicted in Figure~\ref{fig:computationgraph}.

\subsection{Skeleton Refinement (Inner Search)}
\label{sec:refinement}
Skeleton refinement refers to the process of evaluating stream instances in the computation graph in order to produce assignments of a skeleton's continuous parameters.
These sampling operations can fail if there are no feasible outputs conditioned on its inputs. For instance, as shown in Figure~\ref{fig:computationgraph}, if we sample a particular grasp $?g_1$ for the pick action in the first step of the plan, there may be no feasible inverse kinematics solution $?q_2$ for the subsequent placement action. Therefore, we have to be able to backtrack to reconsider the choice of grasp. This is common in sampling-based TAMP approaches, and there are many possible strategies that may be used. In this work, we use a simple strategy which backtracks to the first action upon reaching a dead-end. See Algorithm \ref{lst:refinement}. It can be shown that this strategy is probabilistically complete if the streams produce samples with replacement. 

\begin{figure}
\centering
\begin{lstlisting}[label=lst:refinement,caption={Skeleton Refinement},basicstyle=\small]
def $\sample(\hlplan, N_{max}, \stats)$
  for d := 0 to $N_{max}$
    $\groundings := \emptyset$
    for $a \in \hlplan$
        for $\stream \in$ a.streams
            $\bar o$ = next($\stream$)
            $\stats[\stream]$.attempts++
            if $\bar o$ = null
                break # stream failed
            $\stats[\stream]$.success++
            $\groundings := \groundings \cup \{\bar o\}$
        if all streams successful
            record $\groundings$ as partial grounding of $a$
        elif any partial grounding $\groundings'$ of $a$ exists
            $\groundings := \groundings'$
        else break # deadend
    return $\groundings$ if all actions grounded else null
\end{lstlisting}
\vspace{-2em}
\end{figure}

\subsection{Incorporating Feedback}
\label{sec:feedback}
If skeleton refinement fails, this means that we were unable to find feasible assignments for one or more of the parameters of some action(s) in the skeleton. We would therefore like to modify the $\priority$ value for failed actions, so that the next iteration of Algorithm \ref{lst:lazysp} may avoid them. By maintaining statistics about the success or failure of stream instances in the computation graph of the skeleton, we can empirically estimate the probability of successfully sampling a feasible value for each parameter in the plan and identify bottlenecks.



Since each action includes one or more parameters, we define the estimate of feasibility for an action in a given state $\feasibility(a|s)$ as: 
\begin{equation}
    \feasibility(a|s) := \min_{\stream \in a.streams} \frac{N^\stream_{success} + 1}{N^\stream_{attempts} + 1}
\end{equation}
Note that before we have sampled a particular action's parameters, this definition leads to an estimate of $\feasibility(a|s) = 1$, meaning that we assume initially that all actions are feasible.

We incorporate feasibility estimates into the $\priority$ function in each iteration after a failed refinement. When $\priority$ is defined as in A*, these estimates replace the unit cost associated to each action in the cost-to-come $g(\node)$ - we detail this in section \ref{sec:astarfeedback}. Similarly, we describe how these feasibility estimates are incorporated into the policy when using the LevinTS implementation of $\priority$ in \ref{sec:levintsfeedback}.

\subsubsection{Computation Graph Keys}
As described in section \ref{sec:skeletonsearch}, each node in our search tree maintains a computation graph that defines the sequence of streams which produce each of the parameters/objects in the plan skeleton. Note that different skeletons may include objects with different identifiers which have the same computation graph. For example, any skeleton which includes an action that picks up an object is going to have a parameter corresponding to a grasp of that object.  If we find that one such parameter has low feasibility (i.e. the sequence of streams that should produce it repeatedly fail), then this information should carry over to other plans which include ``similar'' parameters. Therefore, we define the concept of a CG \textit{key}. If two objects share a CG key, this means that barring a renaming of variables, they have identical computation graphs.

\subsubsection{Feedback in A* Priority}
\label{sec:astarfeedback}
When using the A* priority function for $\priority$, we define the cost of an action in our plan as 
$c(a|s) = \frac{1}{\feasibility(a|s)}$.
This means that in the first iteration of planning, when the feasibility of actions is optimistic, the planner uses unit costs for all actions. Similarly, all actions whose computation graphs have never been encountered in refinement will have unit cost. On the other hand, actions whose parameters have failed to be refined will have their costs increased, and thus be deprioritized. This is akin to a relaxation of the binary edge evaluation in \cite{lazysp}.

\subsubsection{Feedback in LevinTS Priority}
\label{sec:levintsfeedback}
If during the course of sampling a candidate plan, we find that $a_2$ is infeasible, then we would like to \textbf{decrease} the policy's probability of taking that action in the next iteration of algorithm \ref{lst:lazysp}. So, given an edge feasibility function $\feasibility(a|s) \to [0,1]$ we define $$\bar{\pol}(a|\state, \goalset) = \frac{\pol(a|\state, \goalset)\feasibility(a|s)}{\sum_{a'}\pol(a'|\state, \goalset)\feasibility(a'|s)}$$

\noindent We use $\bar{\pol}(a|\state, \goalset)$ to define $\bar{\pol}(\node)$ in the same way as described in Equation \ref{eq:levin}.
Note that prior to obtaining empirical estimates for $\feasibility$, we have $\bar{\pol} = \pol$. Actions which are found to be infeasible are deprioritized in subsequent iterations.

\subsection{Architecture and Training of the Skeleton Search Policy}


As described in section \ref{sec:guidance}, in order to guide the skeleton search (using the LevinTS priority function), we require a policy $\pol(a|\state, \goalset)$ that assigns a probability distribution over applicable actions in a given state. One challenge here is that, since we are performing a search in the space of plan skeletons, we only have access to the low-level state in the initial scene. This is because the actions (i.e. logical transitions) that we consider during our skeleton search have parameters (e.g. motions, poses, etc) which are left unspecified. For instance, a plan skeleton which optimistically places an object on the table will not specify the precise grasp used, or the precise final pose of the object on the table.
We would like to learn a policy $\pol(a|\hat \state,\goalset)$ where $\hat \state = \langle \initialstate , \hlplan \rangle$ describes the low-level initial state, and the logical partial skeleton, and $\goalset$ describes the set of desired facts in the goal. 

\subsubsection{State and Goal Representation}
In this work, we consider policies parametrized by Graph Neural Networks \cite{battaglia2018relational}. An illustration of the end-to-end architecture is shown in Figure \ref{fig:model_flow}. We encode a  state $\state$ and goal $\goalset$ as a graph, with nodes representing objects (e.g. table2, block1, robot) and edges between them representing facts which hold (e.g. block1 is on table2) in $\state$ or $\goalset$, following prior work \cite{silver2020planning, khodeir2021learning}. Node features encode the type of object, the precise 3D pose (if unchanged from $\initialstate$), and size of the object. We use Graph Attention Networks (GAT) \cite{velickovic2018graph, brody2022how} to produce contextual embeddings for each of the objects. 

\subsubsection{Action Representation}
The set of applicable actions $\actions(\state)$ at a given state comprise the domain of the probability distribution which should be predicted by the policy. Each action consists of the name of an operator (e.g. pick/place/stack/unstack) encoded using a 1-hot vector of fixed size, as well as a tuple of discrete parameters, whose encodings are obtained from the final layer of the GAT. We employ a second attention network which allows each of these actions to attend to every object in the state, and produce an embedding which is then passed to a simple multilayer perceptron and softmax layer to produce the final probability distribution over actions.


\subsubsection{Training and Data Collection}
We use behavior cloning to train the policy from demonstrations on the set of training problems $\{\initialstate^{(i)}, \goalset^{(i)}\}_{i=1}^N$. We generate these using Algorithm \ref{lst:lazysp} with the A* priority function described in \ref{sec:guidance} and \ref{sec:astarfeedback}.

Note that the returned action sequences $\hlplan(\groundings)^{(i)}$ will have all of their continuous parameters fully specified. In order to train the policy for skeleton search, we extract the high level actions $a^{(i)}_{1:T^{(i)}}$ from the returned trajectory, and use the known high level transition function to extract the sequence of high level states $\state^{(i)}_{0:T^{(i - 1)}}$. 

We then construct a dataset consisting of goal, state and action tuples $\{ \langle \goalset^{(i)}, \state^{(i)}_{j - 1}, a_{j} \rangle \}_{i=1,j=1}^{i=N,j=T^{(i)}}$ and train our models to minimize the cross-entropy loss between the demonstration and predictions.

\section{Experimental Results}

Our experiments are designed to shed light on the following research questions:
(Q1) How well does $\lifted$ perform when used with off-the-shelf domain-agnostic search heuristic?
(Q2) How effective is the learned policy at guiding the skeleton search?
(Q3) Which of the policy-guided search variants described in \ref{sec:Approach} best incorporate the learned policy?

\label{sec:Experiments}
\subsection{Problem Types}
Evaluation of $\lifted$ was conducted across five problem types involving a 7DoF robot arm. 
Problems are divided into five categories which share a domain definition, but present different challenges to the planner. Example scenes are shown in Figure \ref{fig:experiments}. There are two types of blocks (not distinguished logically): blocks (shorter) and blockers (taller). 

In \underline{Stacking}, blocks are arranged randomly in each scene, and the goal is to assemble them into specific towers. 
In \underline{Sorting}, the goal is to move colored blocks to the table with the corresponding color. Blockers may need to be moved if they obstruct a plan, but must be returned to their original tables. Test problems involved up to 10 blocks and 10 blockers.
In \underline{Random}, blocks need to be stacked or rearranged, and blockers may obstruct actions. \underline{Clutter} problems are similar to Random, but contain twice as many blockers, and initial positions are sampled using ordered Poisson-Disc Sampling so that there is a higher chance of obstructions. 
Finally, in \underline{Distractors}, blocks need to be stacked or rearranged in the presence of "distractor" objects which are placed on a separate table. Unlike the blockers in other problem types, distractors do not appear in the goal, and never need to be interacted with. This tests the planner's ability to ignore irrelevant objects. 

For each problem type, we randomly generate 100 instances which are used to train the model and 100 held-out test instances which we use for evaluation. We do not train the model on any Distractors problems, but instead reserve these just for testing. \textit{The training instances are drawn from distributions with few objects/goals so that the baseline is able to solve the majority of them within the timeout. We sample more challenging instances from a different distribution for testing to evaluate the model's ability to generalize to harder problems with more objects than those seen during training.} Initial placements and goals are randomly generated in each problem, so that even problems of the same type with the same number of objects will have different solutions.

All experiments are conducted using 2 cores of an Intel Broadwell processor with an 8GB memory limit. All methods use a 90 second planning timeout, so we report the proportion of problems solved within the timeout, as well as average planning times for solved instances.

\subsection{Results and Discussion}

\begin{wraptable}{l}{4cm}
 \vspace{-.3em}
\scriptsize
\centering
\tabcolsep=0.06cm
\begin{tabular}{lcc}
\toprule
 & adaptive & $\lifted$($h_{add}$) \\
Problem     &          &       \\
\midrule
Random           &    72.00 $\pm$ 1.00 & \textbf{91.80 $\pm$ 1.10} \\
Clutter          &    55.40 $\pm$ 1.52 & \textbf{61.00 $\pm$ 1.41} \\
Stacking         &    61.40 $\pm$ 0.89 & \textbf{88.00 $\pm$ 2.35} \\
Sorting          &    \textbf{77.20 $\pm$ 4.15} & 68.00 $\pm$ 1.87 \\
Distractors &    86.20 $\pm$ 1.10 & \textbf{99.80 $\pm$ 0.45} \\
\bottomrule
\end{tabular}
\caption{}
\label{tab:hadd}
\vspace{-2.2em}
\end{wraptable}
To address (Q1), we evaluate $\lifted$($h_{add}$) which uses the popular domain-agnostic heuristic ``$h_{add}$'' \cite{BONET20015} using the A* priority function. This also establishes a baseline with which to compare the learned policy guided version of $\lifted$ for (Q2). In the table on the left, we compare $\lifted$($h_{add}$) to adaptive \cite{garrett2020pddlstream} in terms of the percentage of test problems solved during the allotted time. We find that across 4 out of 5 problem types, $\lifted$($h_{add}$) solves significantly more problems within the allotted planning time. Adaptive only outperforms $\lifted$($h_{add}$) on sorting problems. We found that this to be the result of increased node expansion time due to the larger number of objects/goals in those problems. As adaptive relies on the efficient implementation of FastDownward, it is able to handle this more effectively. We attribute the improvement on the remaining 4 problem types to the feedback process described in section $\ref{sec:astarfeedback}$ enabling a more efficient search for a feasible plan skeleton.

To address (Q2/Q3) we evaluate 3 variants of policy-guided search from our framework. The first two (i.e. $\lifted(\text{beam}_1)$, and $\lifted(\text{beam}_{10})$) are instances of beam search with beam widths of $W=1$ and $W=10$ respectively. The third (i.e. $\lifted(\text{bfs})$), is an instance of best-first-search. All of these variants use the LevinTS priority function from equation \ref{eq:levints}.

\begin{figure}
\begin{subfigure}[b]{\linewidth}
    \centering
    \includegraphics[width=7cm,height=4cm]{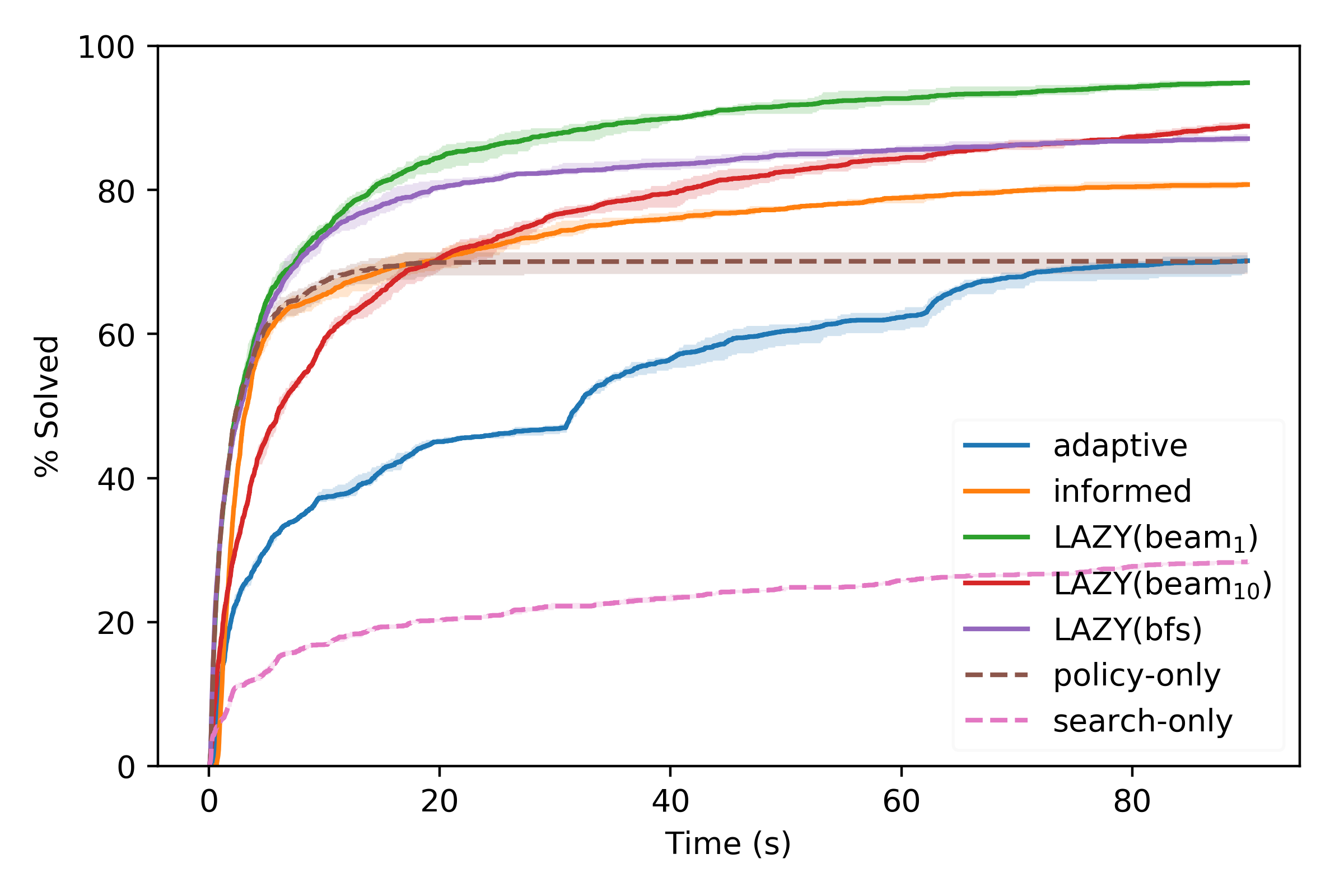}
\end{subfigure}
     \begin{subtable}[b]{\linewidth}
        \scriptsize
        \centering
        \tabcolsep=0.06cm

 
\begin{tabular}{c|ccccc}
\toprule
{}	& \text{Random}	& \text{Clutter}	& \text{Stacking}	& \text{Sorting}	& \text{Distractors}           \\ \midrule
\text{informed}	& {89.8 $\pm$ 0.84}	& {64.0 $\pm$ 1.00}	& {73.0 $\pm$ 1.00}	& {59.0 $\pm$ 1.82}	& \textbf{99.8} $\pm$ 0.45 \\ 
\text{$\lifted(\text{beam}_1)$}	& {\textbf{99.0} $\pm$ 0.00}	& {\textbf{90.8} $\pm$ 1.10}	& {\textbf{91.6} $\pm$ 1.82}	& {\textbf{97.2} $\pm$ 0.84}	& \textbf{100} $\pm$ 0.00  \\ 
\text{$\lifted(\text{beam}_{10})$}	& \textbf{97.8} $\pm$ 0.45	& \text{82.0} $\pm$ 1.41	& \textbf{93.6} $\pm$ 0.55	& 78.4 $\pm$ 3.36	& 96.6 $\pm$ 1.14                           \\ 
\text{$\lifted(\text{bfs})$}	& \textbf{98.0} $\pm$ 0.00	& \text{79.2} $\pm$ 1.64	& \textbf{93.6} $\pm$ 0.55	& 68.8 $\pm$ 1.30	& \textbf{100} $\pm$ 0.00  \\ \midrule
\text{policy-only}	& {83.8 $\pm$ 1.92}	& {43.4 $\pm$ 0.55}	& {76.2 $\pm$ 2.39}	& {51.4 $\pm$ 0.89}	& 98.6 $\pm$ 2.19                           \\ 
\text{search-only}                      \	& 29.0 $\pm$ 0.00	& 24.2 $\pm$ 0.084	& 45.0 $\pm$ 0.00	& 12.0 $\pm$ 0.00	& 32.4 $\pm$ 0.55   \\ \bottomrule                       
\end{tabular}

\end{subtable}

\caption{Figure shows solve rate as a function of planning time. Table reports percentage of problems solved within 90 second timeout. All variants of $\lifted$ use the LevinTS priority function with the learned policy. We report the mean and standard deviation across 5 random seeds for each method.}
\label{tab:learning_solved}
\vspace{-1.9em}
\end{figure}

In Figure \ref{tab:learning_solved}, we compare these variants of our approach to INFORMED \cite{khodeir2021learning}, a prior work which uses learned models to prioritize the inclusion of stream instances according to their predicted relevance to the planning problem.
While all of these methods outperform both non-learning baselines from table \ref{tab:hadd}, we find that $\lifted(\text{beam}_1)$ is consistently the highest performer, solving an average of 96\% of test problems in the allotted time. This suggests that the learned policy is effective at guiding search. However, in general, we expect that the answer to (Q3) will depend on the quality of the policy. 

We also report the performance of two ablations of our method. The first, ``policy-only'' simply uses the learned policy greedily to find a single plan skeleton which it tries to refine for the remainder of the time. The second, ``search-only'' uses the lazy search with feedback framework without a guidance policy.  The results show that, although the policy is effective at guiding search, it is not sufficiently good as to do away with search altogether, and benefits greatly from the overarching framework. Similarly, the relatively poor performance of ``search-only'' demonstrates that both components contribute significantly to the overall success of $\lifted(\text{beam}_1)$.

\begin{figure}[ht]
    \centering
    \includegraphics[height=4cm]{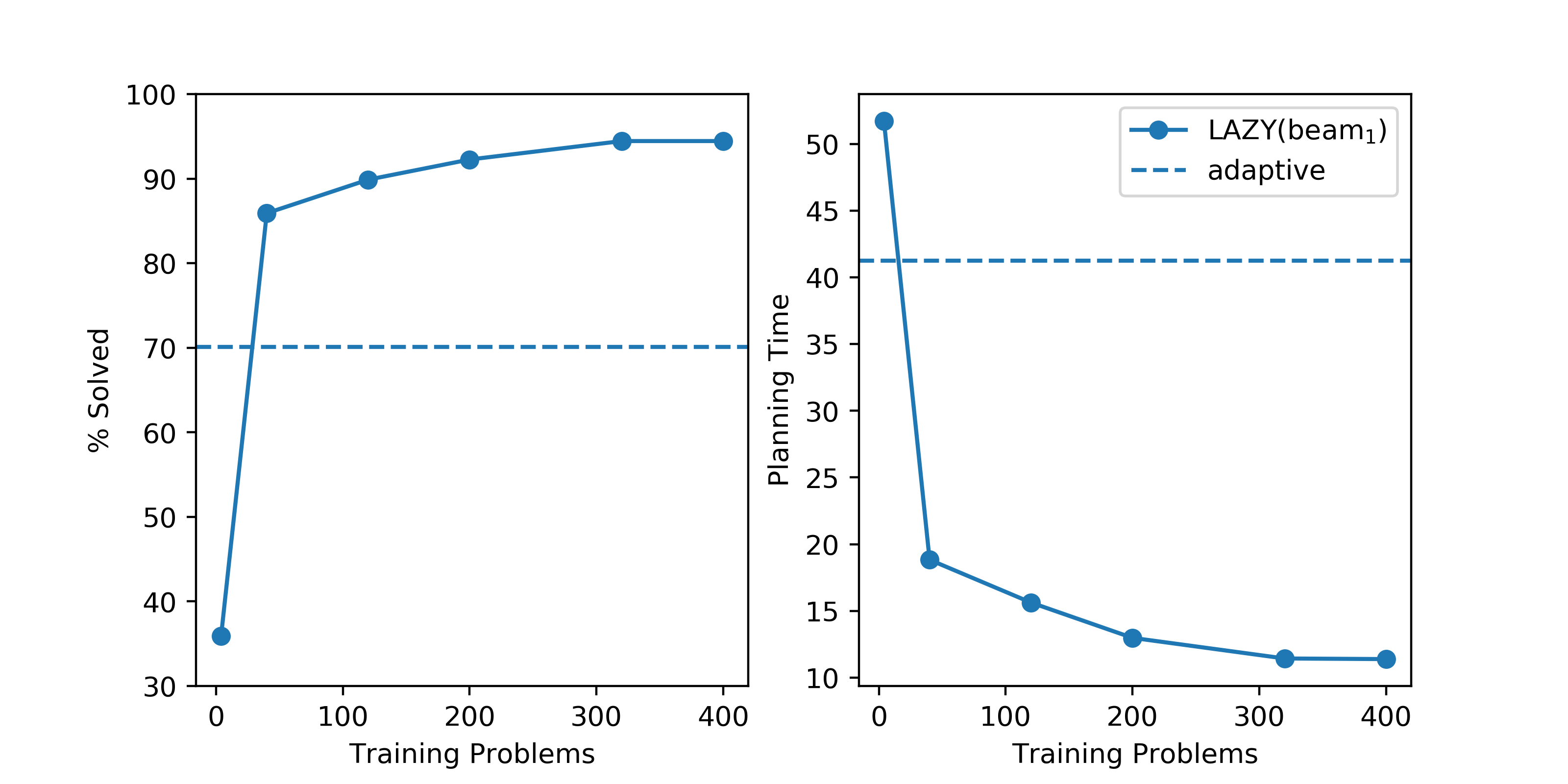}

    \caption[Planning performance of $\lifted$ as a function of training set size]{Left: Percentage of test problems solved by $\lifted(\text{beam}_1)$ as a function of the training set size. Right: Average planning time as a function of training set size.}
    \label{tab:lazy_train_v_solve}
\vspace{-2em}
\end{figure}

In order to shed light on the effect of training set size on the performance of the planner, we trained policies on increasing subsets of the full training set, and evaluated their performance on the test set. We report the percentage of problems solved within the 90 second timeout, as well as the average planning time as a function of the training set size. We find that $\lifted(\text{beam}_1)$ outperforms baselines with only 50 training examples, and continues to improve on both metrics as more training data is used.


\section{Related Work}
\label{sec:Related}

\textbf{Integrated task and motion planning.} There is a vast literature on the problem of integrating the geometric reasoning required by motion planning with the symbolic reasoning that is necessary for planning to achieve abstract goals; see \cite{garrett2020integrated} for a detailed taxonomy. 
Our work builds on the PDDLStream \cite{garrett2020pddlstream} formalism, which we introduced in detail in the background section. 
Several algorithms for PDDLStream problems have since been proposed, including \cite{extendedtreesearch} which uses Monte-Carlo Tree Search to efficiently search for a low-cost solution. 
PDDLStream has also been used to facilitate belief-space planning in partially observed environments \cite{9196681}.

There is a long history of prior research, including~\cite{cambon, plaku2010sampling} combining symbolic planners with complete geometric planners.
The need for selecting correct hierarchical abstractions for symbolic planning and favoring feasibility and real-time results over optimality was emphasized in~\cite{hierarchical_tamp_now}. Logic Geometric Programming combined symbolic planning and trajectory optimization~\cite{toussaint2015logic, toussaint_stable_modes}, even for dynamic physical motions involving tool use, while~\cite{z3_tamp} integrated sampling procedures with SAT solvers. The idea of incorporating refinement failures to bias symbolic search away from infeasible actions was explored in \cite{z3_tamp, srivastava2014combined, lagriffoul2016combining}.
\nocite{hauser2009integrating}

\textbf{Learning for TAMP}. Motivated by the success of learning in the context of robotics, recent work has sought to combine the ability of TAMP systems to plan for novel temporally extended goals with learning methods. 
Under this umbrella, there are: methods which learn continuous action samplers for capabilities that may be difficult to engineer (e.g. pouring) \cite{wang2020learning, kim2018guiding}, those which learn the symbolic representations with which to plan \cite{silver2021learning, loula2020learning, diehl2021automated}, those that integrate perception learning and scene understanding into TAMP~\cite{deep_pddl_tamp_policies, Zhu2020HierarchicalPF}, and those which attempt to learn search guidance from experience \cite{driess2020deep, driess2020deeph, xu2022accelerating, kim2019learning}. A closely related work in this context is \cite{beomjoon2020learning}, where the learned search guidance takes the form of a symbolic action-value function which is also parametrized by a GNN.

\section{Conclusion} 
\label{sec:Conclusion}
In this work, we proposed bi-level lazy search guided by learned goal-conditioned policies as a method for solving TAMP problems expressed using the PDDLStream formalism. We evaluated this approach experimentally against existing solvers, including one prior work which uses learned models, and demonstrated significant improvements in planning times and solve rates across a range of unseen manipulation problems using a 7DoF robot arm.



\bibliographystyle{IEEEtran}
\bibliography{ICRA2023}

\newpage
\appendix

\subsection{Pseudocode}
We provide pseudocode for the standard search routines employed by $\lifted$ in our experiments in the interest of completeness. 
\begin{figure}[h!]
\begin{lstlisting}[numbers=left,numberstyle=\tiny,label=lst:bfs,caption={Best-first search.}]
def BFS($\rootnode$, $\goalset$, $\priority$)
  $\closed := \emptyset$
  $\fringe := \{\rootnode\}$
  while $\fringe \neq \emptyset$
    $\node := \argmin_{\node\in\fringe} \priority(\node)$
    $\fringe := \fringe \setminus\{\node\}$
    $\state := \trans(\node)$
    if $\state \in \goalset$
      return $\node$
    if $\state \in \closed$
      continue
    $\fringe := \fringe \cup \children(\node)$
    $\closed := \closed \cup \{\state\}$  (* \label{lst:levints:cutstop} *)
  return null
\end{lstlisting}
\begin{lstlisting}[label=lst:beam,caption={Beam Search}]
def $\text{Beam}_W$($\rootnode$, $\goalset$, $\priority$)
  $\closed := \emptyset$
  $\fringe := \{\rootnode\}$
  while $\fringe \neq \emptyset$
    $\fringe_W := \emptyset$
    while $|\fringe_W| < W$ and $|\fringe| > 0$
        $\node = \argmin_{\node\in\fringe} \priority(\node)$
        $\fringe := \fringe \setminus \{\node\} $
        $\fringe_W := \fringe \cup \{\node\} $
    $\fringe := \emptyset$
    for $\node \in \fringe_W$
        $\fringe := \fringe \setminus\{\node\}$
        $\state := \trans(\node)$
        if $\state \in \goalset$
            return $\node$
        if $\state \in \closed$
            continue
        $\fringe := \fringe \cup \children(\node)$
        $\closed := \closed \cup \{\state\}$ 

  return null
            
\end{lstlisting}




\end{figure}
\end{document}